# On Detection Reliability for Small Object Segmentation

**Olivier Rukundo**

Department of Electronic and Computer Engineering
University of Limerick; e-mail: olivier.rukundo@ul.ie

**Abstract:** This study investigates whether predicted bounding boxes are reliable for determining where small-object segmentation should be performed without risking false negatives outside those boxes. The investigation leverages adenovirus particles as small objects, YOLO26x for detection, YOLO26x-Sem for segmentation, and Mask R-CNN for both detection and segmentation. Segmentation-model predictions were matched to ground-truth adenovirus masks or reference objects using their centroids, with ground-truth mask centroids preferred over ground-truth bounding-box centroids as references for matching. Matched predicted-mask centroids were considered as inside at least one detector-predicted bounding box or outside all such boxes, while ground-truth masks without matched predictions were recorded as having no matched segmentation. For the YOLO26 models, 34 reference objects (47.89%) had matched segmentation centroids inside at least one predicted box, 8 (11.27%) had centroids outside all predicted boxes, and 29 (40.85%) had no matched predicted segmentation. For the Mask R-CNN model, the corresponding counts were 69 (97.18%), 0 (0.00%), and 2 (2.82%). These preliminary experiments demonstrated that restricting segmentation to detector-predicted bounding boxes risks excluding matched segmentation predictions outside those boxes, potentially introducing false negatives.

## 1. Introduction

Small object segmentation can become very challenging when neighbouring objects must be distinguished from one another and/or from surrounding noisy and complex background or structures [1], [2]. Generally, segmentation tasks are about delineating the boundary of objects of interest while detection tasks are about locating objects of interest in a given image or video [41-50]. Segmentation models may extend predicted masks beyond object boundaries or assign foreground labels to surrounding structures [56-60]. The YOLO26s-Sem and YOLO26m-Sem examples in Figure 1 show that possibility where predicted foreground extends beyond the annotated adenovirus boundaries and includes regions without corresponding ground-truth masks. Even if, detection and segmentation can also be combined within a common framework [3], [4], [5], using predicted bounding boxes to determine where segmentation should occur implies a dependence on the detector, which can be useful only if those boxes do not exclude relevant objects that the segmentor could otherwise identify. In this context, it is important to investigate the reliability of predicted bounding boxes before relying on them to determine where segmentation should occur. One relevant study examines the influence of bounding-box inputs on segmentation [6]. Its authors introduced bounding-box robustness evaluation of promptable segmentation, leveraging adversarial bounding-box prompts to minimize or maximize segmentation error while adhering to naturalness constraints [6]. Other previous studies have examined detector-error propagation, bounding-box prompt sensitivity, and methods for recovering segmentation beyond detector-proposed regions [61–65]. For example, in [61], [62], the authors introduced frameworks leveraging the Segment Anything Model (SAM)'s capability to interpret bounding boxes as prompts, thus depending on the object detector's outcome (e.g., if an object of interest is missed by the detector, SAM can never output a mask for it; if the box is a false positive, SAM generally outputs a false mask).

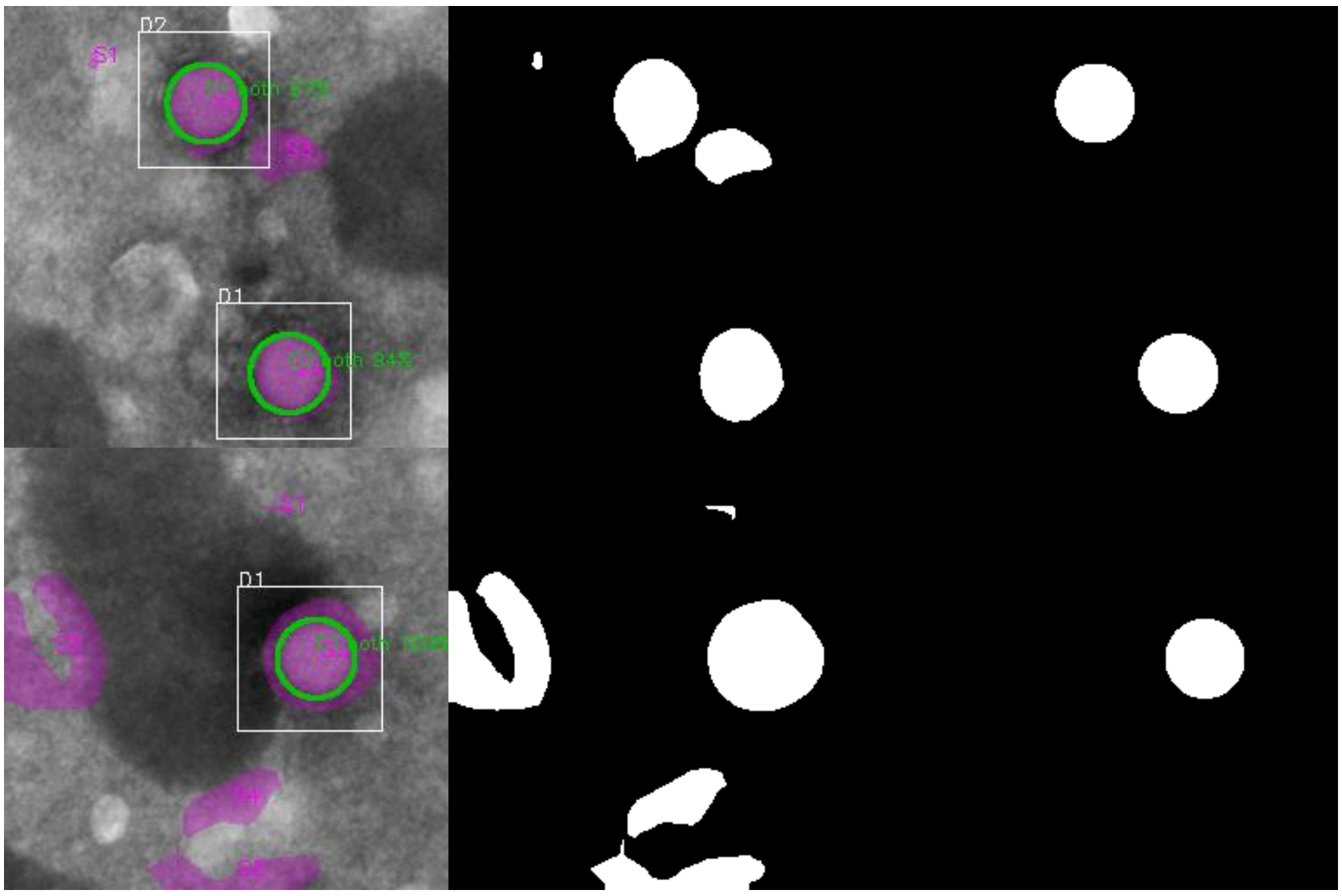

Figure 1: Segmentation errors from YOLO26s-Sem (top) and YOLO26m-Sem (bottom). Left: prediction overlays; middle: predicted masks; right: ground-truth masks.

In [63], the authors presented a probabilistic framework for reasoning about regions, objects, and their attributes, such as object class, location, and spatial extent, by leveraging a conditional random field defined on pixels, segments, and objects. In [64], the authors proposed a novel segmentation approach rather than a fixed-size bounding-box-based segmentation approach that leverages a rectangle as a soft constraint based on a Euclidean distance map.

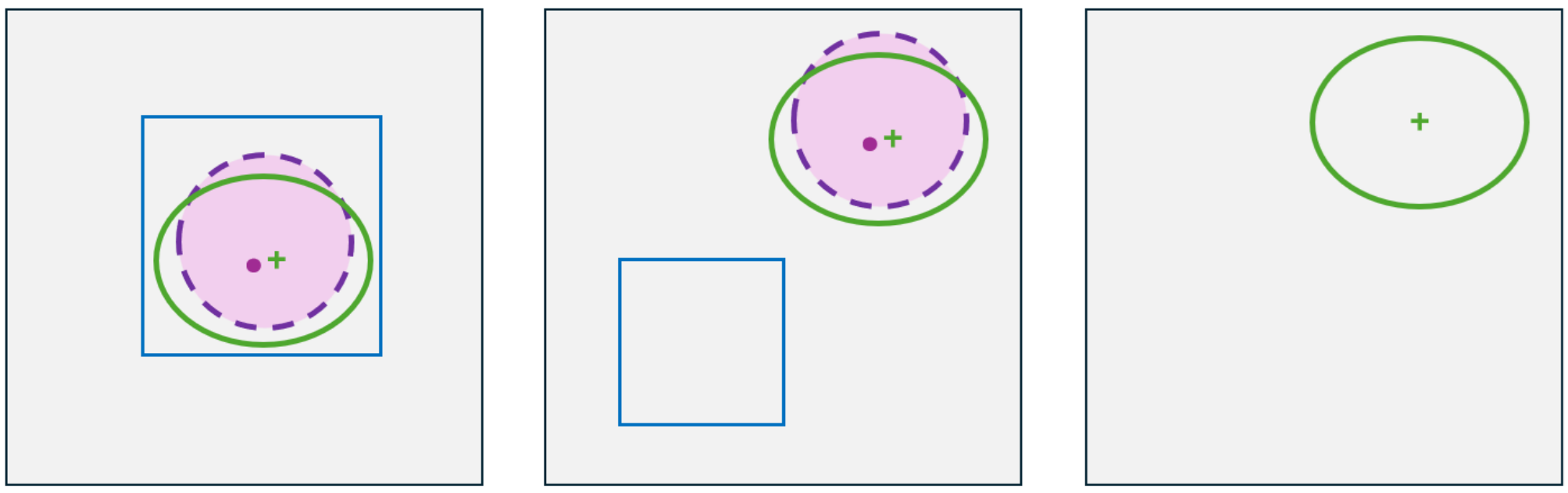

(a) (b) (c)

Figure 2: Schematic illustration of (a) an inside match, (b) an outside match, and (c) no match. Green outlines represent ground-truth masks, purple dashed outlines represent matched predicted masks, and blue rectangles represent predicted bounding boxes. Green crosses and purple dots indicate ground-truth and predicted-mask centroids, respectively.

In [65], the authors proposed Tiny-YOLOSAM, leveraging YOLOv12-based box prompts and supplementing uncovered regions only where YOLO-guided masks provide no coverage. Our investigation differs from all of these in terms of its purpose. This present study instead quantifies the spatial correspondence between predicted segmentation masks and detection bounding boxes using ground-truth matching and three outcomes—inside a predicted box, outside all predicted boxes, and no matched segmentation—to investigate the risk of restricting segmentation to detector-predicted regions or bounding boxes. Figure 2 shows the three cases considered in this investigation. After matching predicted segmentation masks to ground-truth masks using their centroids, matched predicted-mask centroids are classified as inside at least one predicted bounding box (Figure 2-a) or outside all predicted bounding boxes (Figure 2-b). Ground-truth masks without matched predicted segmentation are recorded separately (Figure 2-c). It is understandable that detection precision describes the correctness of accepted detections [51-55], but does not establish that every relevant object has received a region for segmentation. Similarly, an overall segmentation score does not directly describe the spatial correspondence between independently predicted masks and detection bounding boxes. Adenovirus analysis in Transmission Electron Microscopy (TEM) provides the experimental setting for this investigation of small object segmentation [7-10]. Also, the present investigation also connects these two earlier investigations or studies [9],[10], by evaluating detection and segmentation models on SSP dataset derived from the original adenovirus class of the TEM virus dataset [14].

## 2. Materials and Methods

### *2.1. Reference Annotation Consistency Assessment*

Before evaluating model predictions, consistency between the shared reference or ground-truth detection and segmentation annotations was checked across the test split. Individual foreground objects were identified in the reference masks using eight-connected component analysis. Their bounding boxes were expanded by five pixels on each side and clipped to the image boundaries. Corresponding detection-label and segmentation-mask files were compared, and object counts were checked. The centre of each reference bounding box and the centroid of each foreground mask were calculated using Eq. (1). The mask centroid follows the standard image-moment definition [11].

$$b_i = \left(\frac{x_i^{min}+x_i^{max}}{2}, \frac{y_i^{min}+y_i^{max}}{2}\right),\ c_j = \left(\frac{1}{|M_j|}\sum_{(x,y)\in M_j} x\,, \frac{1}{|M_j|}\sum_{(x,y)\in M_j} y\right) \quad (1)$$

where $b_i$ is is the bounding-box centre, $M_j$ is the foreground-pixel set of mask object $j$, and $|M_j|$ is its number of foreground pixels. Within each image, one-to-one correspondence was established by minimizing the total Euclidean distance between bounding-box centres and mask centroids.

$$\pi^* = \underset{\pi \in \Pi}{arg\,min} \sum_{i=1}^{n} \left\|b_i - c_{\pi(i)}\right\|_2 \quad (2)$$

where $n$ is the common number of reference boxes and mask objects in the image, and $\Pi$ is the set of one-to-one assignments. Eq. (2) was solved using SciPy's *linear_sum_assignment*, which implements a modified Jonker–Volgenant algorithm [12, 13]. After matching within each image, all matched pairs were pooled across the test split and indexed by $k = 1, \ldots, N$. Centre-coordinate agreement was defined as the percentage of pairs whose Euclidean centre distance did not exceed the diagnostic tolerance $\tau = 5$ pixels.

$$P_{agree} = \frac{100}{N}\sum_{k=1}^{N} 1\left[\left\|b_k - c_k\right\|_2 \le \tau\right],\quad P_{disagree} = 100 - P_{agree} \quad (3)$$

where $1[.]$ denotes the indicator function, which equals one when the condition is satisfied and zero otherwise. The five-pixel tolerance was selected for this diagnostic comparison and was distinct from the five-pixel margin used to construct the reference boxes. Separately, mask-centroid containment was defined as the percentage of matched mask centroids located inside their corresponding reference bounding boxes.

$$P_{inside} = \frac{100}{N}\sum_{k=1}^{N} 1[c_k \in B_k],\quad P_{outside} = 100 - P_{inside} \quad (4)$$

where $B_k$ denotes the corresponding reference bounding box, including its boundary. Both percentages used the total number of matched reference pairs as the denominator. Centre-coordinate agreement and mask-centroid containment, calculated using Eq. (3) and Eq. (4) are presented in Figure 3 and Figure 4, respectively.

### 2.2. Detection and Segmentation Models

The models comprise five YOLO26 object detectors (*n, s, m, l, and x*), five corresponding YOLO26-Sem semantic segmentors, and Mask R-CNN with a ResNet50–FPN backbone. YOLO26 detectors output bounding boxes and object-confidence scores, while YOLO26-Sem models output foreground-probability maps. Mask R-CNN provides bounding boxes, confidence scores, and instance masks [3]. The union of its retained instance masks represents the foreground for pixel-level evaluation. The centroid-containment analysis comprises the YOLO26x detector, YOLO26x-Sem segmentor, and Mask R-CNN. Detection-confidence and foreground-probability thresholds are 0.25 and 0.50, respectively. Note that all these models were trained on the MEDION workstation equipped with a single NVIDIA GeForce RTX 3070 GPU and an 11th-generation Intel Core i7-11700F processor operating at 2.50 GHz, with eight cores and 16 logical processors.

### 2.3. Training datasets

This dataset contains adenovirus image patches derived from the adenovirus class of the TEM Virus Dataset introduced in [8]. The patches were created using the Semi-Automatic Stride-Independent Patching (SSP) tool developed in [10]. The SSP dataset contains 145 image patch pairs for detection and segmentation [14]. The training, validation, and test splits contain 87, 26, and 32 patches, respectively [14]. Detection and segmentation models shared the same image splits. Each detection image has a corresponding bounding-box annotation file, while each segmentation image has a corresponding binary reference mask. The test split contains 71 reference objects. Detection annotations followed YOLO format: class identifier, normalized horizontal and vertical centre coordinates, width, and height. The adenovirus class identifier was 0. Binary segmentation masks represented adenovirus foreground with pixel value 255 and background with pixel value 0. Each eight-connected foreground component defined an individual reference object. Its detection bounding box included a five-pixel margin on each side, subject to image boundaries. Mask R-CNN targets comprised individual component masks, corresponding expanded boxes, and foreground class labels. These targets derive from the same binary masks used for semantic segmentation, with box coordinates checked against the YOLO detection labels.

### 2.4. Augmentation operations

To increase training-data variability and reduce overfitting, augmentation of the training split was done using geometric transformations [10], [15-18]. Such geometric transformations mainly included random rotation between −180° and 180°, horizontal and vertical translation of up to 10% of the corresponding image dimension, uniform scaling by a factor between 0.8 and 1.2, and independent horizontal and vertical flipping, each with a probability of 0.5 [9]. Most geometric transformations rely on extra-pixel and non-extra-pixel interpolation algorithms [19–40] for training split data processing, respectively. However, recent research has demonstrated benefits of extra-pixel interpolation for training split data geometric transformations to improve deep learning models training outcomes [15], [16].

### 2.5. Hyperparameter settings

Training configurations specified a maximum of 300 epochs and an early-stopping patience of 100 epochs. YOLO26 detectors used a mini-batch size of 2 and the MuSGD optimizer. YOLO26-Sem used automatic optimizer selection, with mini-batch sizes of 6 for the n, s, and m variants and 1 for the l and x variants. Mask R-CNN uses a mini-batch size of 2 and SGD with an initial learning rate of 0.005, momentum of 0.9, weight decay of 0.0005, and cosine learning-rate scheduling. Unspecified training options retain their implementation defaults. Manual hyperparameter adjustment follows an observation-based criterion informed by previous studies [15–18]: no further adjustments were made if the monitored training metric reached 0.90 within the first 30 epochs.

## 3. Results

Experimental results are provided into three subsections related to reference annotation consistency, detection and segmentation performance on the SSP test split and predicted-mask centroid containment and qualitative examination of predictions.

### 3.1. Reference Annotation Consistency Results

Across 32 test patches containing 71 reference objects, 69 bounding-box centre–mask centroid pairs (97.18%) had a Euclidean distance of ≤5 pixels, while two pairs (2.82%) exceeded this tolerance (Figure 3). All 71 mask centroids (100%) were located inside their corresponding reference bounding boxes (Figure 4).

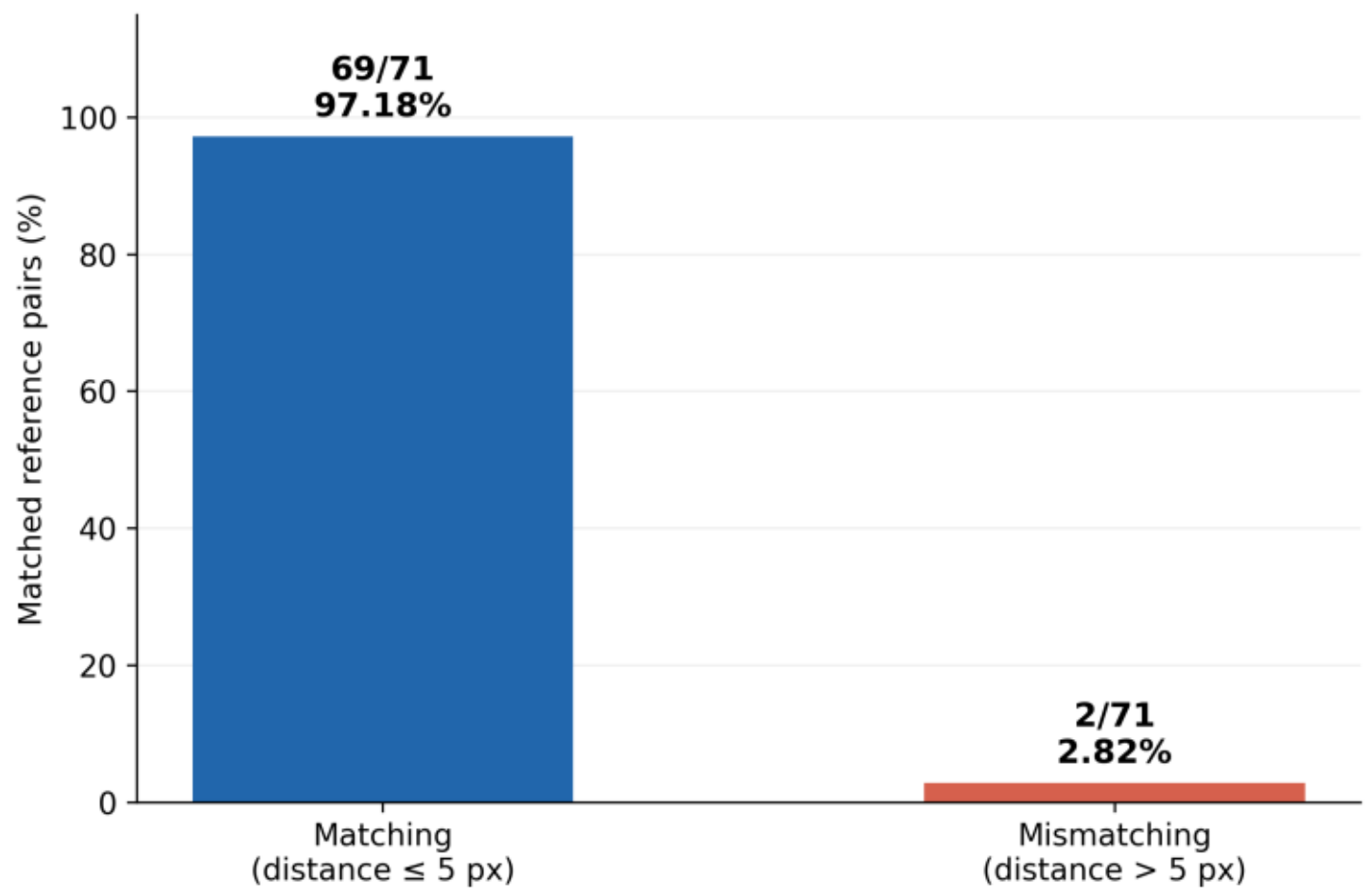

Figure 3: Reference center-coordinate agreement. Euclidean centre-distance tolerance: 5 pixels.

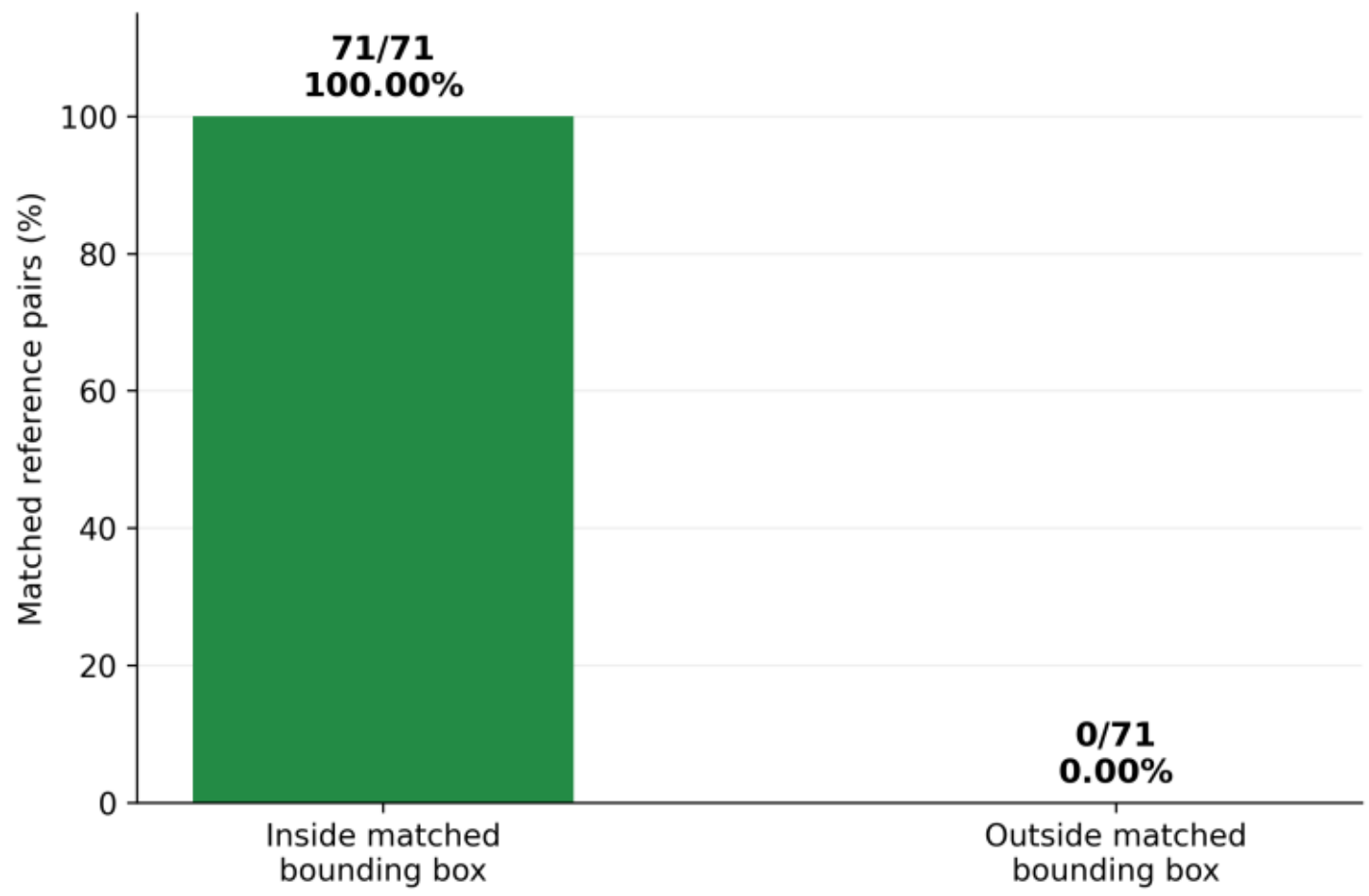

Figure 4: Reference mask-centroid containment. Containment is assessed against the corresponding reference bounding box.

### 3.2. Detection and Segmentation Performance

All five YOLO26 model-size variants were evaluated to mesure the detection (Table 1) and segmentation (Table 2) performance on the SSP test set. YOLO26x and YOLO26x-Sem were selected for the subsequent centroid-containment analysis because they generally achieved the highest mAP@0.50:0.95 and foreground IoU, respectively, among the evaluated YOLO26 variants.

Table 1: Detection models performance on the SSP test split

| Model | Precision | Recall | mAP@0.50 | mAP@0.50:0.95 |
|---|---|---|---|---|
| YOLO26n | 0.9210 | 0.8169 | 0.8978 | 0.5320 |
| YOLO26s | 0.8479 | 0.7069 | 0.7738 | 0.4149 |
| YOLO26m | 0.8411 | 0.8201 | 0.8892 | 0.4753 |
| YOLO26l | 0.8219 | 0.7324 | 0.8490 | 0.5110 |
| YOLO26x | 0.9456 | 0.8028 | 0.9075 | 0.5480 |
| Mask R-CNN | 0.8519 | 0.9718 | — | — |

**Note**: Mask R-CNN precision and recall were calculated using centre-distance matching, whereas YOLO26 metrics came from its detector evaluation procedure. "—" indicates an unavailable result.

Table 2: Segmentation models performance on the SSP test split. Mask R-CNN instance masks were combined into a single foreground mask for pixel-level evaluation

| Model | IoU | mIoU | Dice | Pixel Precision | Pixel Pecall |
|---|---|---|---|---|---|
| YOLO26n-Sem | 0.6196 | 0.7680 | 0.7651 | 0.6391 | 0.9529 |
| YOLO26s-Sem | 0.6137 | 0.7639 | 0.7606 | 0.6321 | 0.9548 |
| YOLO26m-Sem | 0.6035 | 0.7564 | 0.7527 | 0.6173 | 0.9642 |
| YOLO26l-Sem | 0.6145 | 0.7784 | 0.7612 | 0.8817 | 0.6697 |
| YOLO26x-Sem | 0.7144 | 0.8325 | 0.8334 | 0.7959 | 0.8747 |
| Mask R-CNN | 0.7254 | 0.8367 | 0.8409 | 0.7468 | 0.9620 |

*3.3. Predicted-Mask Centroid Containment*

Figure 5 shows the obtained results. For YOLO26x/YOLO26x-Sem, 34 objects (47.89%) had matched segmentation centroids inside at least one predicted bounding box, whereas eight (11.27%) had centroids outside all predicted boxes.

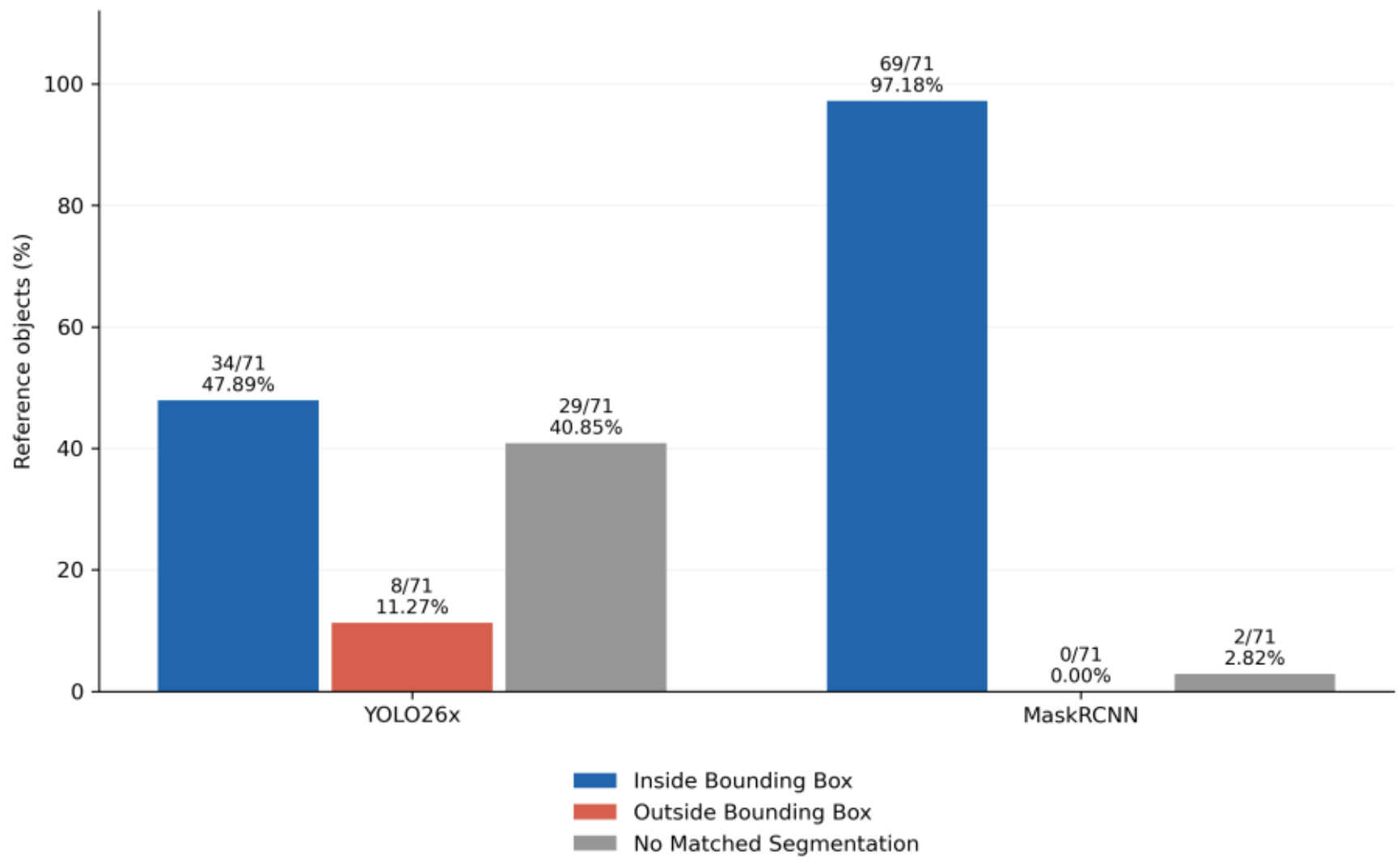

Figure 5: Results achieved by YOLO26 and MASK-R-CNN models.

The remaining 29 objects (40.85%) had no matched predicted segmentation under the specified matching criterion. For Mask R-CNN, 69 objects (97.18%) had matched segmentation centroids inside predicted boxes, none were classified as outside, and two (2.82%) had no matched segmentation.

*3.4. Qualitative Examination of Predictions*

Figure 6 shows representative examples of the three outcomes reported in Section 3.3. The example of a matched segmentation centroids inside at least one predicted bounding box (or an inside predicted box or an inside match). The example of a centroid outside the predicted box (or outside predicted boxes or an outside match) and the example showing a no matched predicted segmentation (or no matched segmentation or no match). Here, the important distinction is that "no matched segmentation" does not mean "no segmentation." In Figure 6-c, purple predictions may exist but fail the matching criterion for the ground-truth objects marked by yellow crosses. Note that, Figure 2-c simply illustrates that category without showing unmatched predictions.

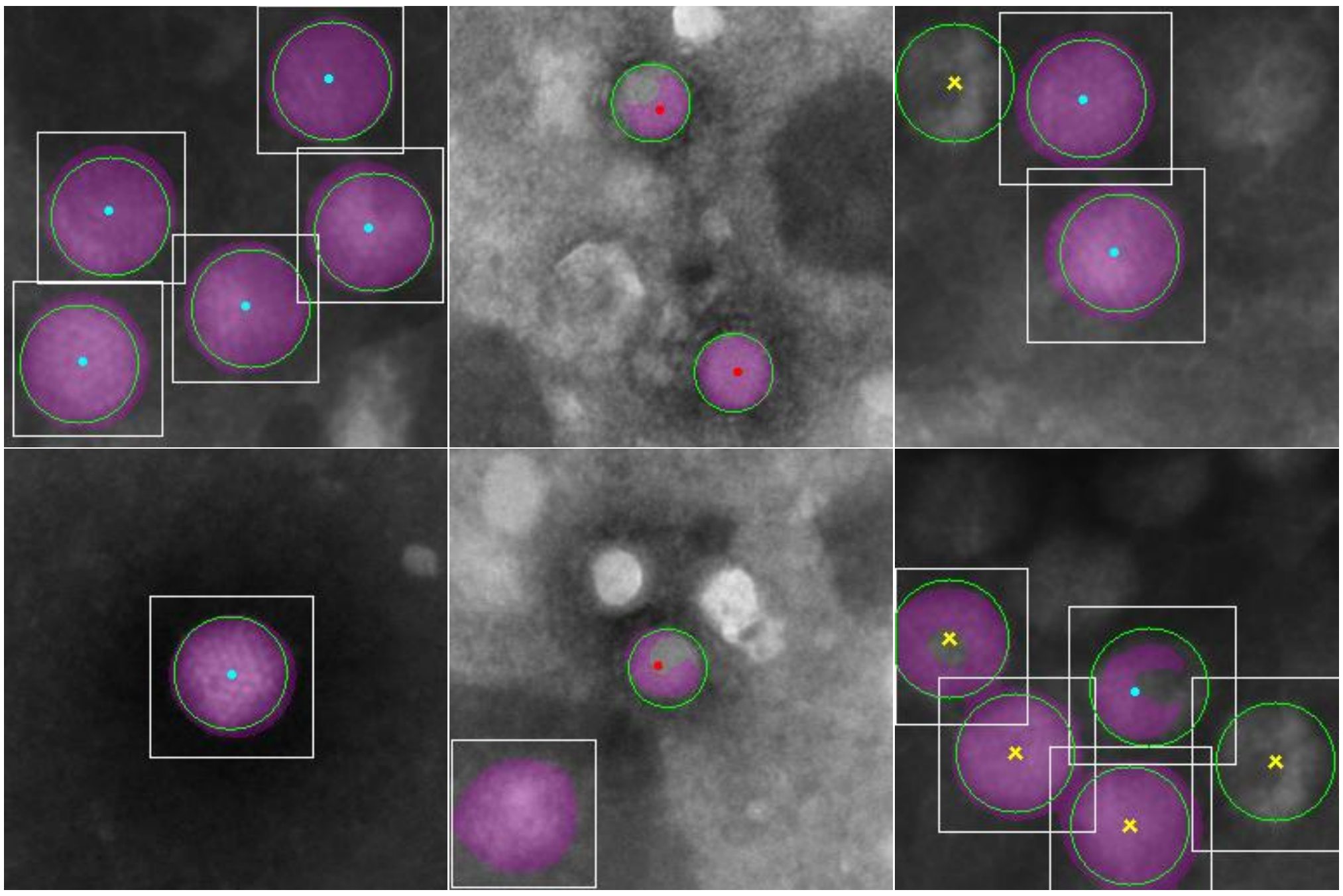

Figure 6: (left): Matched predicted-mask centroids inside at least one predicted bounding box. (middle): Matched predicted-mask centroids outside all predicted bounding boxes. (right): Ground-truth objects without matched predicted segmentation masks, marked by yellow crosses.

## 4. Discussion

The reference annotation consistency assessment established the reference-level correspondence needed for this study. The detection and segmentation model performance on the SSP dataset demonstrated why YOLO26x and YOLO26x-Sem were chosen for further analysis. Understandably, direct detection comparisons with Mask R-CNN require harmonized matching protocols. Detection reliability findings demonstrated that restricting segmentation to detector-predicted regions risks excluding matched segmentation predictions, potentially introducing false negatives. However, in all 69 matched Mask R-CNN mask centroids fell within predicted boxes, but this containment is constrained by its associated box–mask outputs [3]. It does not establish detection completeness. The two ground-truth objects without matched segmentation indicate a segmentation-matching outcome, rather than necessarily a detection failure. Objects missed by both branches cannot appear as outside-box segmentations. Note that, our analysis examines whether matched predicted-mask centroids fall outside all detector-predicted bounding boxes. It assesses spatial correspondence rather than the performance of segmentation rerun within detected boxes. Matching-sensitivity analysis, and evaluation on broader datasets are needed before generalizing the findings.

## 5. Conclusions

This study presented preliminary findings on the reliability of predicted bounding boxes for determining where small-object segmentation should occur. The experiments demonstrated that matched predicted-mask centroids were not consistently located inside detector-predicted bounding boxes. This indicates that restricting segmentation to detector-predicted regions risks excluding matched segmentation predictions, potentially introducing false negatives.

**Supplementary Materials:** Source codes developed can be obtained upon request.

**Author Contributions:** O.R. performed all aspects of this work and has read and approved this version of the manuscript.

**Funding:** This research received no external funding.

**Data Availability Statement:** The used SSP datasets are publicly available on Zenodo: https://doi.org/10.5281/zenodo.23402752

**Acknowledgments:** The author would like to thank you for constructive comments.

**Conflicts of Interest:** The authors declare no conflicts of interest.

## References

1. Schmidt, U.; Weigert, M.; Broaddus, C.; Myers, G. Cell Detection with Star-Convex Polygons. In Medical Image Computing and Computer Assisted Intervention—MICCAI 2018; 2018; pp. 265–273
2. Stringer, C.; Wang, T.; Michaelos, M.; Pachitariu, M. Cellpose: A Generalist Algorithm for Cellular Segmentation. Nature Methods 2021, 18, 100–106
3. He, K.M., Gkioxari, G., Dollár, P., Girshick, R., "Mask R-CNN," 2017 IEEE International Conference on Computer Vision (ICCV), Venice, Italy, 2017, pp. 2980-2988
4. Kirillov, A.; et al. Segment Anything. In Proceedings of the IEEE/CVF International Conference on Computer Vision (ICCV); 2023; pp. 4015–4026
5. Marks, M.; et al. CellSAM: A Foundation Model for Cell Segmentation. Nature Methods 2025, 22, 2585–2593
6. Moskalenko, A.; et al. BREPS: Bounding-Box Robustness Evaluation of Promptable Segmentation. Proceedings of the AAAI Conference on Artificial Intelligence 2026, 40(10), 8107–8115
7. Rukundo, O., Behanova, A., De Feo, R. et al. Convolutional Neural Networks for Automatic Detection of Intact Adenovirus from TEM Imaging with Debris, Broken and Artefacts Particles, arXiv preprint arXiv: 2310.1963, 2023
8. Matuszewski, D. J.; Sintorn, I.-M. TEM Virus Images: Benchmark Dataset and Deep Learning Classification. Computer Methods and Programs in Biomedicine 2021, 209, 106318
9. Rukundo, O., Toward Optimal Adenovirus Detection Using YOLO26, arXiv preprint arXiv:2607.17799, 2026
10. Rukundo, O., Stride Independent Patching for Deep Learning, arXiv preprint arXiv:2610.12216, 2026
11. OpenCV, "Contour Features," OpenCV Documentation. [Online]. Available: https://docs.opencv.org/4.x/dd/d49/tutorial_py_contour_features.html. Accessed: Oct. 22, 2026
12. SciPy, "linear_sum_assignment," SciPy Documentation. [Online]. Available: https://docs.scipy.org/doc/scipy/reference/generated/scipy.optimize.linear_sum_assignment.html. Accessed: Oct. 22, 2026
13. Crouse, D. F. On Implementing 2D Rectangular Assignment Algorithms. IEEE Transactions on Aerospace and Electronic Systems 2016, 52(4), 1679–1696
14. Rukundo, O. (2026). SSP Adenovirus Dataset: TEM Image Patches with Detection and Segmentation Annotations [Dataset]. Zenodo. https://doi.org/10.5281/zenodo.23402752
15. Rukundo, O. Evaluation of Extra Pixel Interpolation with Mask Processing for Medical Image Segmentation with Deep Learning. SIViP 18, 7703–7710, 2024
16. Rukundo, O., Beyond Nearest Neighbor Interpolation in Data Augmentation, SIViP, 2026
17. Rukundo, O. Effect of the regularization hyperparameter on deep learning-based segmentation in LGE-MRI. In Proceedings of the SPIE/COS Photonics Asia, Nantong, China, 10–20 October 2021; Volume 11897
18. Rukundo, O. Effects of Image Size on Deep Learning. Electronics 2023, 12, 985
19. Rukundo, O. Non-extra pixel interpolation. Int. J. Image Graph. 2020, 20, 2050031
20. Rukundo, O. Optimal Methods Research on Grayscale Image Interpolation; China National Knowledge Infrastructure CNKI, TP391.41: Beijing, China, 2012
21. Shangguan, J.-T.; Li, Y.-L.; Wang, Y.-G.; Li, H.-L. Fast Algorithm of Modified Cubic Convolution Interpolation. In Proceedings of the 2011 4th International Congress on Image and Signal Processing, Shanghai, China; 2011; pp. 1072–1075
22. Bobokhonov, A.; Xuramov, L.; Rashidov, A. Evaluation of the Effectiveness of Interpolation Methods in the Process of Image Size Standardization. In Proceedings of the 2025 International Russian Automation Conference (RusAutoCon), Sochi, Russian Federation; 2025; pp. 165–170
23. Nayak, R.; Patra, D. Image Interpolation Using Adaptive P-Spline. In Proceedings of the 2015 Annual IEEE India Conference (INDICON), New Delhi, India; 2015; pp. 1–6
24. Rukundo, O. Evaluation of Rounding Functions in Nearest Neighbour Interpolation. Int. J. Comput. Methods 2021, 18, 2150024
25. Rukundo, O.; Schmidt, S.E.; Von Ramm, O.T. Software implementation of optimized bicubic interpolated scan conversion in echocardiography. arXiv 2020, arXiv:2005.11269
26. Wang, W.; Xiong, Z.; Li, H.; Li, D. Comparison Discrete Smooth Interpolation and Traditional Spatial Interpolation. In Proceedings of the 2011 International Conference on Remote Sensing, Environment and Transportation Engineering, Nanjing, China; 2011; pp. 185–187

27. Rukundo, O.; Schmidt, S. Stochastic Rounding for Image Interpolation and Scan Conversion. Int. J. Adv. Comput. Sci. Appl. 2022, 13, 13–22
28. Rukundo, O.; Huang, M.; Cao, H. Optimization of Bilinear Interpolation Based on Ant Colony Algorithm. In Proceedings of the 2nd International Conference Electrical and Electronics Engineering, Macao, China, 1–2 December 2011; pp. 571–580
29. Tian Yan, Liu Bin and Li Tao, "A Local Image Interpolation Method Based On Gradient Analysis," 2005 International Conference on Neural Networks and Brain, Beijing, 2005, pp. 1202-1205
30. Rukundo, O.; Wu, K.; Cao, H. Image Interpolation Based on The Pixel Value Corresponding to The Smallest Absolute Difference. In Proceedings of the 4th International Workshop on Advanced Computational Intelligence, Wuhan, China, 19–21 October 2011; pp. 434–437
31. K. K. Delibasis and A. Kechriniotis, "A New Formula for Bivariate Hermite Interpolation on Variable Step Grids and Its Application to Image Interpolation," in IEEE Transactions on Image Processing, vol. 23, no. 7, pp. 2892-2904, July 2014
32. Rukundo, O. Effects of improved-floor function on the accuracy of bilinear interpolation algorithm. *Comput. Inf. Sci*. 2015,8,1-25
33. Rukundo, O.; Maharaj, B. Optimization of Image Interpolation based on Nearest Neighbour Algorithm. In Proceedings of the 9th International Conference on Computer Vision Theory and Applications, Lisbon, Portugal, 5–8 January 2014; pp. 641–647
34. Chi-Wah Kok; Wing-Shan Tam, "Nonadaptive Interpolation," in Digital Image Interpolation in Matlab , IEEE, 2019, pp.91-122
35. Rukundo, O. Normalized weighting schemes for image interpolation algorithms. Appl. Sci. 2023, 13, 1741
36. Hiep Quang Luong, P. de Smet and W. Philips, "Image interpolation using constrained adaptive contrast enhancement techniques," IEEE International Conference on Image Processing 2005, Genova, Italy, 2005, pp. II-998
37. Rukundo, O.; Cao, H. Advances on image interpolation based on ant colony algorithm. SpringerPlus 2016, 5, 403
38. Rukundo, O.; Schmidt, S. Extrapolation for image interpolation. In Proceedings of the SPIE/COS Photonics Asia, Beijing, China, 11–13 October 2018; Volume 10817
39. Liu, X.; Zhao, D.; Xiong, R.; Ma, S.; Gao, W. Image Interpolation via Regularized Local Linear Regression. In Proceedings of the 28th Picture Coding Symposium, Nagoya, Japan; 2010; pp. 118–121
40. Gadermayr, M.; et al. Which Factors Influence Success of Unconstrained Interpolation for Augmentation in Multiple Instance Learning? In Proceedings of the 2026 IEEE 39th International Symposium on Computer-Based Medical Systems (CBMS), Limassol, Cyprus; 2026; pp. 43–48
41. Minaee, S.; Boykov, Y.; Porikli, F.; Plaza, A.; Kehtarnavaz, N.; Terzopoulos, D. Image Segmentation Using Deep Learning: A Survey. IEEE Transactions on Pattern Analysis and Machine Intelligence 2022, 44(7), 3523–3542
42. Sharma, R.; Saqib, M.; Lin, C. T.; Blumenstein, M. A Survey on Object Instance Segmentation. SN Computer Science 2022, 3, 499
43. Pham, D. L.; Xu, C.; Prince, J. L. Current Methods in Medical Image Segmentation. Annual Review of Biomedical Engineering 2000, 2, 315–337
44. Zosso, D.; An, J.; Stevick, J.; Takaki, N.; Weiss, M.; Slaughter, L. S.; Cao, H. H.; Weiss, P. S.; Bertozzi, A. L. Image Segmentation with Dynamic Artifacts Detection and Bias Correction. Inverse Problems and Imaging 2017, 11(3), 577–600
45. Ren, Y.; Kong, A. W. K.; Jiao, L. A Survey on Image and Video Cosegmentation: Methods, Challenges and Analyses. Pattern Recognition 2020, 103, 107297
46. Liu, L.; Ouyang, W.; Wang, X.; Fieguth, P.; Chen, J.; Liu, X.; Pietikäinen, M. Deep Learning for Generic Object Detection: A Survey. International Journal of Computer Vision 2020, 128, 261–318
47. Zhao, Z.-Q.; Zheng, P.; Xu, S.-T.; Wu, X. Object Detection With Deep Learning: A Review. IEEE Transactions on Neural Networks and Learning Systems 2019, 30(11), 3212–3232
48. Zou, Z.; Chen, K.; Shi, Z.; Guo, Y.; Ye, J. Object Detection in 20 Years: A Survey. Proceedings of the IEEE 2023, 111(3), 257–276
49. Cheng, G.; Han, J. A Survey on Object Detection in Optical Remote Sensing Images. ISPRS Journal of Photogrammetry and Remote Sensing 2016, 117, 11–28
50. Cazzato, D.; Cimarelli, C.; Sanchez-Lopez, J. L.; Voos, H.; Leo, M. A Survey of Computer Vision Methods for 2D Object Detection from Unmanned Aerial Vehicles. Journal of Imaging 2020, 6(8), 78
51. Padilla, R.; Passos, W. L.; Dias, T. L. B.; Netto, S. L.; da Silva, E. A. B. A Comparative Analysis of Object Detection Metrics with a Companion Open-Source Toolkit. Electronics 2021, 10(3), 279
52. Everingham, M.; Van Gool, L.; Williams, C. K. I.; Winn, J.; Zisserman, A. The PASCAL Visual Object Classes (VOC) Challenge. International Journal of Computer Vision 2010, 88, 303–338
53. Kaur, J.; Singh, W. Tools, Techniques, Datasets and Application Areas for Object Detection in an Image: A Review. Multimedia Tools and Applications 2022, 81, 38297–38351
54. Shah, S.; Tembhurne, J. Object Detection Using Convolutional Neural Networks and Transformer-Based Models: A Review. Journal of Electrical Systems and Information Technology 2023, 10, 54
55. Reinke, A.; et al. Understanding Metric-Related Pitfalls in Image Analysis Validation. Nature Methods 2024, 21, 182–194
56. Karimi, D.; Salcudean, S. E. Reducing the Hausdorff Distance in Medical Image Segmentation With Convolutional Neural Networks. IEEE Transactions on Medical Imaging 2020, 39(2), 499–513
57. Hindsholm, A. M.; Cramer, S. P.; Simonsen, H. J.; Frederiksen, J. L.; Andersen, F.; Højgaard, L.; Ladefoged, C. N.; Lindberg, U. Assessment of Artificial Intelligence Automatic Multiple Sclerosis Lesion Delineation Tool for Clinical Use. Clinical Neuroradiology 2022, 32(3), 643–653
58. Kamnitsas, K.; Ledig, C.; Newcombe, V. F. J.; Simpson, J. P.; Kane, A. D.; Menon, D. K.; Rueckert, D.; Glocker, B. Efficient Multi-Scale 3D CNN with Fully Connected CRF for Accurate Brain Lesion Segmentation. Medical Image Analysis 2017, 36, 61–78

59. Valverde, S.; Cabezas, M.; Roura, E.; González-Villà, S.; Pareto, D.; Vilanova, J. C.; Ramió-Torrentà, L.; Rovira, À.; Oliver, A.; Lladó, X. Improving Automated Multiple Sclerosis Lesion Segmentation with a Cascaded 3D Convolutional Neural Network Approach. NeuroImage 2017, 155, 159–168
60. Taghanaki, S. A.; Zheng, Y.; Zhou, S. K.; Georgescu, B.; Sharma, P.; Xu, D.; Comaniciu, D.; Hamarneh, G. Combo Loss: Handling Input and Output Imbalance in Multi-Organ Segmentation. Computerized Medical Imaging and Graphics 2019, 75, 24–33
61. Tyagi, A. K.; Mishra, V.; Prathosh A. P.; Mausam. BoxCell: Leveraging SAM for Cell Segmentation with Box Supervision. Scientific Reports 2025, 15, 42652
62. Jeon, J.; Lee, S.; Kim, S.; Kim, T. Systematic Evaluation of SAM Prompting Strategies for Agricultural Land Extraction with GeoSAM. Geocarto International 2026, 41(1), 2717446
63. Ladický, L.; Sturgess, P.; Alahari, K.; Russell, C.; Torr, P. H. S. What, Where & How Many? Combining Object Detectors and CRFs. In ECCV 2010; pp. 424–437
64. Xu, N.; Price, B.; Cohen, S.; Yang, J.; Huang, T. Deep GrabCut for Object Selection. In BMVC 2017
65. Xu, K.; Wu, S. Tiny-YOLOSAM: Fast Hybrid Image Segmentation, arXiv:2512.22193, 2025